\documentclass[runningheads]{llncs}
\usepackage[T1]{fontenc}
\usepackage{graphicx,verbatim}
\usepackage{multirow}
\usepackage{amsmath}
\usepackage{amssymb}
\usepackage{booktabs}
\usepackage{marvosym}

\begin{document}
%
\title{AGE-MIL: Anchor-Guided Evidence Learning for Patient-Level Prediction}
\titlerunning{AGE-MIL}
%
\author{
Jiawei Niu\inst{1}\textsuperscript{$\dagger$} \and
Jian Chen\inst{2}\textsuperscript{$\dagger$} \and
Di Zhang\inst{1} \and
Junbo Lu\inst{1} \and
Zhangcheng Liao\inst{3} \and
Xuhao Liu\inst{3} \and
Honglin Zhong\inst{3} \and
Mireia Crispin-Ortuzar\inst{2} \and
Chen Li\inst{1} \and
Zeyu Gao\inst{2}\textsuperscript{(\Letter)} \and
Yi Cai\inst{3}\textsuperscript{(\Letter)}
}

%
\authorrunning{J. Niu et al.}
%
\institute{
School of Computer Science and Technology, Xi’an Jiaotong University, Xi’an, China\\
\and
Department of Oncology, University of Cambridge, Cambridge, UK\\
\and
Xiangya School of Medicine, Central South University, Changsha, Hunan, China\\
\email{zg323@cam.ac.uk; cai-yi@csu.edu.cn}
}

  
\maketitle              
\begingroup

\renewcommand\thefootnote{$\dagger$}

\footnotetext{Contributed equally to this work.}

\endgroup

\begin{abstract}
Existing computational pathology methods predominantly operate within whole-slide image (WSI)-level multiple instance learning (MIL) paradigms, while patient-level modeling remains underexplored. 
In routine pathological practice, however, pathologists derive diagnostic and prognostic conclusions by integrating evidence across multiple WSIs rather than relying on any single slide.
This discrepancy creates a fundamental misalignment when patient-level supervision is directly imposed on conventional MIL frameworks, often leading to unstable optimization and degraded predictive reliability.
To address this issue, we propose Anchor-Guided Evidence MIL (AGE-MIL), a weakly supervised framework for patient-level prediction. AGE-MIL constructs a patient-level anchor from slide representations to capture global pathological context and guide the retrieval and integration of diagnostically relevant local patches, enabling robust patient-level modeling. 
Patient-level risk is further modeled as an evidence accumulation process, promoting stable optimization under weak supervision.
AGE-MIL is evaluated on six clinically relevant patient-level prediction tasks from two independent cohorts. Experimental results show that the proposed framework consistently outperforms eight state-of-the-art MIL methods.
Code is available at https://github.com/wodeniua/AGE-MIL.


\end{abstract}
\section{Introduction}

Computational pathology has achieved substantial progress in tumor diagnosis and prognosis prediction~\cite{campanella2019clinical,gao2025smmile}. Due to the gigapixel resolution of whole slide images (WSIs) and the scarcity of fine-grained annotations, multiple instance learning (MIL) has become the prevailing modeling paradigm, where WSIs are treated as bags and patches as instances~\cite{ilse2018abmil,lu2021clam,li2021dsmil,shao2021transmil}. This formulation has been extensively validated in slide-level prediction tasks, establishing MIL as a powerful paradigm for analyzing large-scale pathological data.

\begin{figure}
\centering
\includegraphics[width=\columnwidth]{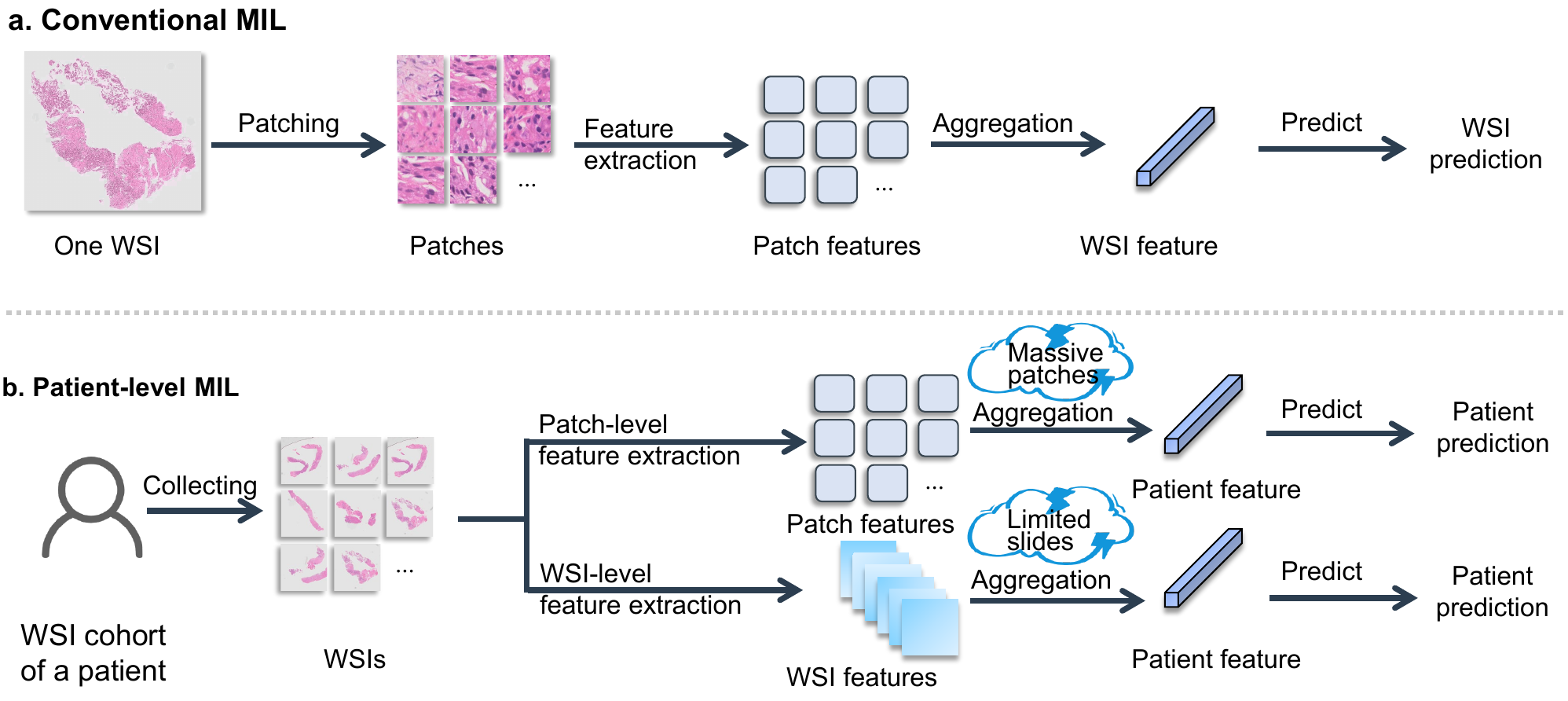}
\caption{Conventional WSI-level MIL vs. patient-level MIL. Patient-level aggregation can be performed either via patch-to-patient or slide-to-patient.}
\label{fig1}
\end{figure}

Clinical decision-making, however, is inherently patient-centric~\cite{courtiol2019centric}. A single patient is typically associated with multiple heterogeneous WSIs, and patient-level conclusions rely on evidence integrated across multiple WSIs rather than any single slide~\cite{campanella2019clinical,diette1997solving}. Directly extending slide-level MIL to patient-level prediction introduces a fundamental mismatch between supervision granularity and evidence distribution~\cite{yao2020cancer_survival}. In particular, patch-to-patient aggregation is computationally prohibitive due to the massive number of instances and the sparse distribution of diagnostic evidence, while also disrupting the patch–slide–patient hierarchy and discarding slide-level structural information as shown in Fig.~\ref{fig1}. 
While recent slide-level foundation models provide semantically consistent and transferable representations~\cite{chief,gigapath}, they primarily operate at the slide level and do not explicitly address patient-level evidence integration. Consequently, directly aggregating slide-level features still risks discarding critical evidence, limiting the ability of existing methods to effectively capture and integrate sparse and distributed patient-level evidence.

To address these challenges, we propose AGE-MIL, an anchor-guided evidence multi-instance learning framework for patient-level prediction. AGE-MIL introduces a patient-level anchor to capture global pathological context and guide the integration of local evidence across multiple slides. Specifically, slide-level representations extracted by the pathology foundation model are aggregated to construct a patient-level anchor with stable semantic characterization. This anchor is then used to guide evidence retrieval from large instance~(patch) pools through patient-aware scoring. The retrieved instances are subsequently integrated via an evidence-aware representation learning module, enabling adaptive refinement of patient representations without requiring additional supervision. Furthermore, patient-level risk is modeled as a progressive evidence accumulation process, promoting stable optimization and robust prediction. 

We evaluate the proposed framework on two large-scale prostate cancer biopsy datasets comprising over 700 patients and approximately 10,000 WSIs from two independent clinical cohorts, covering six clinically relevant patient-level prediction tasks, including lymph node metastasis and prognosis prediction. We further conduct comprehensive patient-level benchmarking against eight state-of-the-art MIL methods. Experimental results show that AGE-MIL improves AUC and accuracy by approximately 2\% on average across the six tasks, establishing state-of-the-art performance.

\section{\textbf{Related Work}}
\subsubsection{\textbf{Multiple Instance Learning.}}
MIL has become a dominant framework in computational pathology~\cite{ilse2018abmil,zhang2025patches}, largely motivated by the gigapixel resolution of WSIs and the lack of fine-grained annotations~\cite{tellez2019neural}. Within this paradigm, WSIs are modeled as bags and image patches as instances, with embedding-level and attention-based aggregation strategies~\cite{ilse2018abmil} widely adopted for slide-level prediction.
However, conventional MIL formulations typically rely on slide-level supervision, implicitly assuming that each bag corresponds to a single WSI~\cite{li2021dsmil,liu2024pamil,niu2025learning}. This assumption becomes problematic in patient-level prediction settings. Assigning patient-level labels to individual slides or instances often introduces supervision ambiguity. Moreover, diagnostically relevant pathological evidence is frequently sparse, creating a structural mismatch with aggregation mechanisms that depend on densely informative instances. Such discrepancies may lead to noisy gradients and the dominance of spurious instances. Although hierarchical MIL variants have been explored~\cite{hou2022h2mil,chen2022hipt,zheng2022graph}, challenges related to supervision noise and evidence sparsity remain insufficiently addressed.

\subsubsection{\textbf{Pathology Foundation Models.}}
Foundation models have emerged as a dominant paradigm in computational pathology, largely driven by large-scale pretraining at the patch and slide levels. These approaches typically combine high-capacity visual encoders with slide-level aggregation mechanisms to learn transferable representations from extensive WSI collections. Representative directions include attention-based aggregation~\cite{chief}, coordinate-aware modeling~\cite{titan,zhang2026care}, long-sequence architectures~\cite{gigapath}, and cross-modal alignment~\cite{prism,tangle,gao2025alpaca}. Pretrained representations have been shown to improve semantic consistency and generalization across downstream tasks. Existing studies primarily focus on encoder design and pretraining strategies, while the role of pretrained features in downstream inference remains less explored. Under weak supervision and patient-level prediction settings, semantically consistent representations naturally enable alternative modeling strategies that emphasize evidence selection and information aggregation rather than instance-level representation learning.

\section{\textbf{Method}}

\begin{figure}[t]
\centering
\includegraphics[width=\columnwidth]{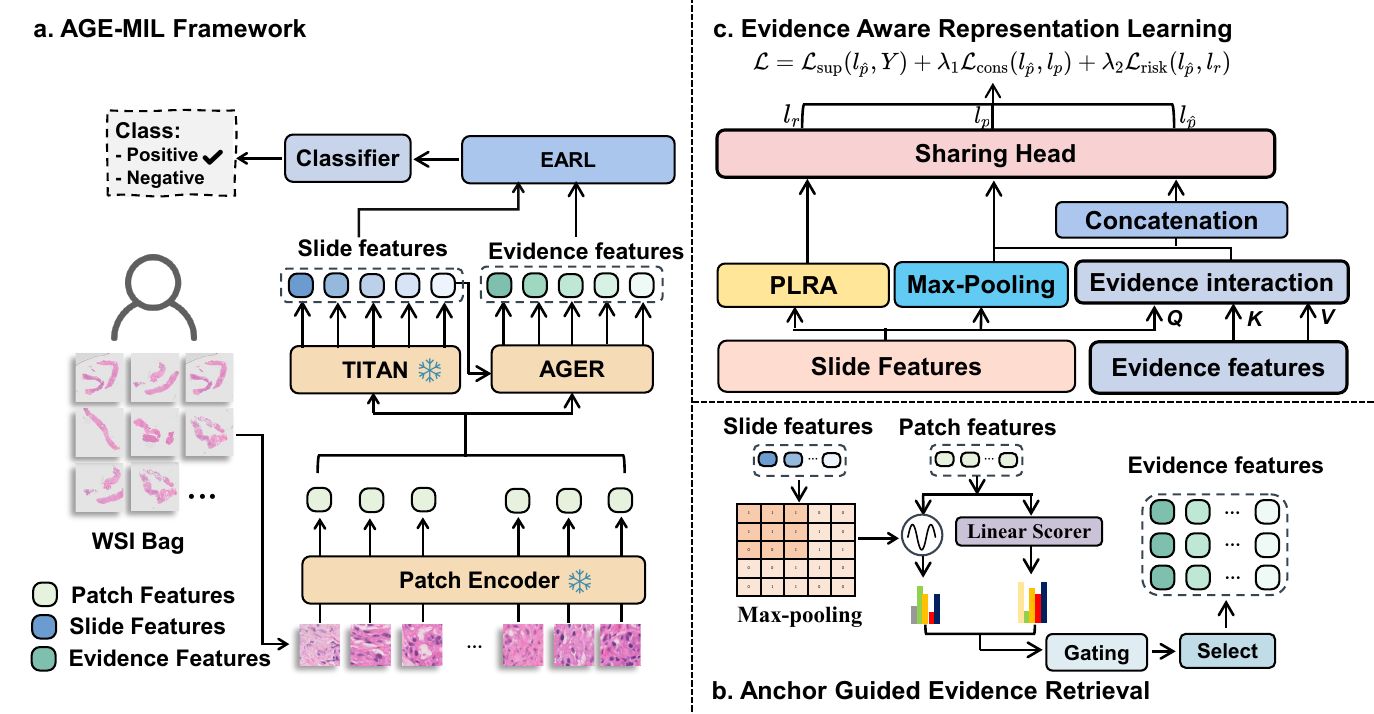}
\caption{Overview of AGE-MIL.
a) AGE-MIL Framework: Patch features are aggregated into slide-level features using TITAN, then processed via AGER for evidence feature retrieval or EARL for patient-level interaction with evidence features.
b) AGER: Guided by the anchor, selects evidence features from a large pool of patches.
c) EARL: Patient-conditioned evidence interaction enhances patient-level modeling.} \label{fig2}
\end{figure}

The overall framework is illustrated in Fig.~\ref{fig2} and consists of three modules: (1) Anchor Guided Evidence Retrieval~(AGER), which identifies diagnostically relevant patches; 
(2) Evidence Aware Representation Learning~(EARL), which integrates patient-level and evidence representations; 
and (3) Patient Level Risk Accumulation~(PLRA), which estimates patient-level risk.

\subsection{Problem Formulation}
Consider a patient-level dataset containing $N$ patients,
denoted as $\mathcal{D} = \{(\mathcal{S}, y)\}$, where $y \in \{0,1\}$
indicates the patient label and $\mathcal{S} = \{S\}$ represents the set
of WSIs associated with a patient. Each slide is decomposed into image
patches, yielding an instance set $\mathcal{X} = \{x \in \mathbb{R}^{D}\}$.

The objective is to learn a mapping $f(\mathcal{S}, \mathcal{X}) \rightarrow \hat{y}$
that predicts the patient label by leveraging both slide-level context and
patch-level evidence. This formulation captures the hierarchical nature of
pathological data, where diagnostically relevant patterns may be sparsely
distributed across slides and instances.

\subsection{Anchor Guided Evidence Retrieval}
For each patient $i$, patch-level features are first extracted using a patch-level foundation model, followed by slide-level feature aggregation using a slide-level foundation model. We infer a patient-level anchor $p_i \in \mathbb{R}^{D_p}$ as a latent global state from the set of slide features. In our implementation, this inference is realized by a max aggregation operator, which serves as a robust estimator of the latent state in small-bag patient settings, where attention-based aggregation may exhibit sensitivity to instance variability. This choice is further supported by prior findings that informative instances often correspond to larger feature magnitudes~\cite{chikontwe2024frmil,lee2021weakly}, making max aggregation a simple yet effective mechanism for capturing salient evidence.

Instead of aggregating all patches within a patient's WSIs, we formulate
evidence selection as a patient-conditioned retrieval process and extract a
compact subset of task-relevant patches as diagnostic evidence. Each patch
embedding $x$ is assigned a retrieval score with two complementary components:
an unconditional term $s^{u} = w_u^{\top} x$ that captures intrinsic instance
saliency, and a conditional term $s^{c} = \langle \phi(x), \psi(p) \rangle / \tau$
that measures anchor--instance compatibility. Here, $\phi(\cdot)$ and $\psi(\cdot)$
are learnable projections and $\tau$ is a temperature parameter. The final score
is defined as:
\begin{equation}
s = (1 - \alpha) s^{u} + \alpha s^{c},
\quad
\alpha = \sigma(g(p)),
\end{equation}
where $\alpha$ adaptively trades off generic saliency and anchor-consistent
relevance. Top-$K$ patches ranked by $\{s\}$ form the evidence set $\mathcal{E}$.

\subsection{Evidence Aware Representation Learning}

To perform patient-conditioned evidence interaction, we employ a cross-attention
operator defined as:
\begin{equation}
\mathrm{Attn}(\mathcal{H}, \mathcal{E})
=
\mathrm{softmax}\left(\frac{\mathcal{H}\mathcal{E}^{\top}}{\sqrt{d}}\right)\mathcal{E},
\end{equation}
where $\mathcal{H}$ denotes the set of slide-level features for a patient, $\mathcal{E}$ represents the retrieved evidence embeddings, and d is the dimension of the evidence feature space. This operation yields context-conditioned evidence representations.

To regulate evidence reliability, we introduce a learnable gating mechanism
applied to each evidence token, $\hat{e} = \gamma \cdot e$, where the gate
$\gamma$ is predicted from the evidence feature itself. The gated evidence
tokens are aggregated via a maxpooling operator to obtain a compact evidence
summary $z^{\mathrm{evi}}$.

Finally, the anchor and evidence representations are jointly integrated. They
are concatenated as \( [p, z^{\mathrm{evi}}] \), followed by a linear
transformation: \( \hat{p} = f_{\mathrm{fuse}}([p, z^{\mathrm{evi}}]) \), where
\( f_{\mathrm{fuse}} \) denotes the fusion mapping.

Gradients are detached from the anchor to
prevent retrieval-induced anchor drift. This design preserves a stable global
patient context while allowing retrieved evidence to refine the final
representation.

\subsection{Patient Level Risk Accumulation}

The patient-level risk logits $r$ are derived from slide-level logits
$\{l_j\}_{j=1}^{m}$ using a Log-Mean-Exp (LME) aggregation:
\begin{equation}
r
=
\log \sum_{j=1}^{m} \exp(l_j)
-
\log m,
\end{equation}

This formulation admits a natural interpretation from a log-evidence
accumulation perspective. Specifically, $\exp(l_j)$ can be understood as a
non-negative evidence contribution associated with slide $j$, reflecting
its relative support for patient-level risk. Under this view, the aggregated
risk corresponds to the logarithm of accumulated evidence across slides.
The normalization term $-\log m$ compensates for variations in slide counts,
thereby reducing bias induced by differing numbers of observations.

Compared with hard max pooling, LME yields a smooth approximation that
retains sensitivity to dominant high-risk slides while preserving stable
gradient behavior. The logarithmic form further improves numerical
stability and is consistent with standard probabilistic interpretations
of logits.

The overall training objective combines supervised learning, representation
consistency, and risk alignment:
\begin{equation}
\mathcal{L}_{\mathrm{sup}} = \mathrm{CE}(l_{\hat{p}}, Y),\;
\mathcal{L}_{\mathrm{cons}} = \mathrm{KL}(l_{\hat{p}}, l_p),\;
\mathcal{L}_{\mathrm{risk}} = T^{2}\,\mathrm{KL}\!\left(\frac{l_r}{T}, \frac{\mathrm{stopgrad}(l_{\hat{p}})}{T}\right),
\end{equation}
where $\mathcal{L}_{\mathrm{sup}}$ denotes the supervised classification
loss, $\mathcal{L}_{\mathrm{cons}}$ enforces predictive consistency between
the evidence-aware representation and the patient anchor, and
$\mathcal{L}_{\mathrm{risk}}$ regularizes the risk branch by aligning it with
the primary prediction. The coefficients $\lambda_1$ and
$\lambda_2$ control the relative contributions of each term.
We use $l_{\hat{p}}$ for the evidence-aware prediction logits, $l_p$ for the
patient anchor logits, $l_r$ for the risk logits, and $Y$ for the patient
label.
Thus, the overall loss is formulated as:
\begin{equation}
\mathcal{L} = \mathcal{L}_{\mathrm{sup}} + \lambda_1\,\mathcal{L}_{\mathrm{cons}} + \lambda_2\,\mathcal{L}_{\mathrm{risk}}.
\end{equation}


\section{Experiments and Results}

\subsubsection{Feature Extraction.}

Both slide-level and patch-level features are extracted at $20\times$ magnification with an input resolution of $512 \times 512$ using the TRIDENT tool~\cite{vaidya2025molecular}. The patch-level features are extracted using CONCH v1.5~\cite{lu2024conch}, and these features are aggregated into slide-level representations using TITAN~\cite{titan}. 

\subsubsection{Dataset.}
We evaluate the proposed framework on two independent patient-level pathology datasets collected from the same medical institution. The datasets consist of biopsy WSIs acquired from routine clinical practice, with patient-level labels defined according to corresponding clinical outcomes.
All data were collected under approval from the Institutional Review Board (IRB) of the institution, and the requirement for informed consent was waived due to the retrospective and anonymized nature of the study.


\textbf{Prostate Metastasis Prediction Dataset:}
This dataset is designed for lymph node metastasis (LNM) risk modeling and is formulated as three independent binary classification tasks, defined according to metastasis patterns: distant metastasis, regional metastasis, and the presence of either distant or regional metastasis. It comprises 379 patients, corresponding to 6,476 WSIs. At the instance level, approximately 1.26M patches are sampled from tissue regions for model training and evidence modeling.

\textbf{Prostate Prognostic Prediction Dataset:}
This dataset is constructed for prognosis modeling associated with androgen receptor signaling inhibitors (ARSIs) and is similarly organized into three independent binary classification tasks. The tasks are defined based on three clinically relevant PSA thresholds (0.2, 2.0, and 4.0 ng/mL), corresponding to biochemical recurrence, disease progression, and high-risk disease, respectively. The dataset contains 306 patients and 3,402 WSIs, from which approximately 1.20M patches are extracted.

For both datasets, patients serve as the primary prediction unit, with
labels defined at the patient level. Each patient may correspond to
multiple WSIs. All slides undergo tissue-aware patch sampling to
construct instance sets for weakly supervised modeling.

\subsubsection{Comparison with MIL methods.}
We compare the proposed method against a diverse set of representative
MIL approaches, including
MeanMIL, MaxMIL, ABMIL~\cite{ilse2018abmil}, CLAM-SB/MB~\cite{lu2021clam}, DSMIL~\cite{li2021dsmil}, TransMIL~\cite{shao2021transmil}, and ILRA~\cite{xiang2023ilra}.To adapt these MIL baseline models to the patient-level prediction scenario, we aggregate the slide features of each patient extracted by TITAN. 

Performance is evaluated using classification accuracy (ACC) and the area
under the receiver operating characteristic curve (AUC). 
To ensure robust and unbiased assessment, we adopt a five-fold
cross-validation on all datasets.

\begin{table}[t]
\caption{Performance comparison on three metastasis prediction tasks.}
\label{tab:LN_results}
\centering
\resizebox{\linewidth}{!}{
\setlength{\tabcolsep}{6pt}  
\renewcommand{\arraystretch}{1.2}  
\begin{tabular}{lcccccc}
\hline
\multirow{2}{*}{Method} 
& \multicolumn{2}{c}{LNM (Any)} 
& \multicolumn{2}{c}{LNM (Distant)}
& \multicolumn{2}{c}{LNM (Region)} \\
\cline{2-7}
& ACC & AUC & ACC & AUC & ACC & AUC \\
\hline
MeanMIL & 78.10$\pm$6.21 & 81.06$\pm$5.71 & 81.02$\pm$1.69 & 80.67$\pm$4.35 & 78.06$\pm$4.35 & 77.44$\pm$10.58 \\
MaxMIL & 77.57$\pm$5.03 & \textbf{82.13$\pm$4.63} & 79.94$\pm$2.58 & 81.62$\pm$3.88 & 76.98$\pm$4.63 & 76.12$\pm$10.68 \\
ABMIL & 78.35$\pm$6.66 & 81.60$\pm$6.62 & 79.95$\pm$0.95 & 80.87$\pm$4.64 & 76.74$\pm$4.91 & 78.68$\pm$10.19 \\
CLAM-SB & 77.04$\pm$5.04 & 81.36$\pm$6.20 & 81.29$\pm$1.28 & 80.85$\pm$5.04 & 76.48$\pm$6.19 & 76.01$\pm$14.65 \\
CLAM-MB & 79.15$\pm$5.81 & 81.67$\pm$5.78 & 81.28$\pm$1.94 & 79.93$\pm$5.52 & 77.26$\pm$4.01 & 78.77$\pm$8.24 \\
DSMIL & 73.09$\pm$2.33 & 80.35$\pm$6.09 & 77.81$\pm$1.55 & 73.81$\pm$9.49 & 73.81$\pm$1.06 & 78.49$\pm$9.70 \\
TransMIL & 78.28$\pm$3.02 & 81.87$\pm$4.02 & 81.56$\pm$2.78 & 80.39$\pm$4.81 & 77.32$\pm$6.54 & 76.73$\pm$10.12 \\
ILRA & 77.29$\pm$4.97 & 80.51$\pm$3.78 & 81.54$\pm$3.99 & 79.42$\pm$6.81 & 74.61$\pm$2.74 & 74.18$\pm$6.62 \\
AGE-MIL & \textbf{80.99$\pm$5.54} & 82.09$\pm$5.12 & \textbf{82.35$\pm$2.27} & \textbf{82.24$\pm$2.30} & \textbf{78.58$\pm$5.42} & \textbf{80.64$\pm$7.73} \\
\hline
\end{tabular}
}
\end{table}

\begin{table}[t]
\caption{Performance comparison on three prognostic prediction tasks.}
\label{tab:ARSI_results}
\centering
\resizebox{\linewidth}{!}{
\setlength{\tabcolsep}{6pt}  
\renewcommand{\arraystretch}{1.2}  
\begin{tabular}{lcccccc}
\hline
\multirow{2}{*}{Method} 
& \multicolumn{2}{c}{ARSI (0.2)} 
& \multicolumn{2}{c}{ARSI (2)}
& \multicolumn{2}{c}{ARSI (4)} \\
\cline{2-7}
& ACC & AUC & ACC & AUC & ACC & AUC \\
\hline
MeanMIL & 66.80$\pm$8.90 & 63.24$\pm$11.77 & 74.80$\pm$1.79 & 56.08$\pm$9.69 & 82.00$\pm$2.45 & 62.58$\pm$8.08 \\
MaxMIL & 66.80$\pm$8.67 & 62.41$\pm$14.27 & 74.40$\pm$1.67 & 62.48$\pm$6.72 & 81.20$\pm$2.28 & 61.93$\pm$11.97 \\
ABMIL & 66.40$\pm$9.32 & 61.68$\pm$13.02 & 74.40$\pm$1.67 & 59.04$\pm$8.83 & \textbf{82.40$\pm$2.19} & 62.48$\pm$10.67 \\
CLAM-SB & \textbf{67.20$\pm$8.44} & 64.11$\pm$10.89 & 75.20$\pm$1.10 & 55.73$\pm$8.41 & 82.00$\pm$1.41 & 58.85$\pm$10.28 \\
CLAM-MB & 66.40$\pm$9.10 & 65.56$\pm$12.63 & 73.60$\pm$1.67 & 60.23$\pm$3.75 & 82.00$\pm$2.45 & 60.17$\pm$10.79 \\
DSMIL & 65.60$\pm$6.39 & 63.31$\pm$9.19 & 73.20$\pm$1.79 & 58.80$\pm$6.38 & 80.40$\pm$0.89 & 60.86$\pm$8.36 \\
TransMIL & 62.80$\pm$5.22 & 63.40$\pm$6.21 & 74.00$\pm$1.41 & 53.02$\pm$10.77 & 81.20$\pm$1.79 & 58.28$\pm$11.50 \\
ILRA & 62.80$\pm$3.35 & 64.46$\pm$7.58 & 75.20$\pm$3.35 & 56.11$\pm$12.26 & 82.00$\pm$1.41 & 62.18$\pm$8.52 \\
AGE-MIL & 66.40$\pm$4.28 & \textbf{67.67$\pm$8.82} & \textbf{75.40$\pm$1.72} & \textbf{62.57$\pm$7.75} & 80.80$\pm$2.15 & \textbf{62.71$\pm$8.32} \\
\hline
\end{tabular}
}
\end{table}

From Table \ref{tab:LN_results} and Table \ref{tab:ARSI_results}, our method consistently outperforms state-of-the-art approaches across both datasets and tasks. In the LN dataset, our method improves AUC by 1-3\% over the best competitor, MaxMIL, while in the ARSI dataset, we achieve a notable AUC improvement of over 4\% for ARSI1 and ARSI2 compared to CLAM-MB. These results demonstrate the superior performance and robustness of our approach across various tasks and datasets.

\subsubsection{Ablational Studies.}

We perform an ablation study to assess the contribution of the proposed components, as reported in Table \ref{tab:ablation}. The analysis considers variants without the evidence concatenation, the conditional similarity retrieval mechanism (Cond-Sim), the stop-gradient strategy, and the introduced loss terms.

The observations indicate that all components positively influence model performance. Excluding the evidence pathway leads to clear degradation across datasets, highlighting the benefit of incorporating retrieved evidence. Removing the conditional similarity mechanism likewise reduces accuracy, suggesting that patient-aware retrieval is important for identifying relevant instances. Notably, disabling the stop-gradient strategy results in inferior performance, implying its role in maintaining stable retrieval behavior. In addition, eliminating the proposed loss terms further weakens the results, reflecting their complementary effects during optimization.

\begin{table}[t]
\caption{Ablation study of the proposed method on three LNM datasets. I: Evidence concatenation, II: Cond-Sim, III:
StopGrad, IV: $\mathcal{L}_{\mathrm{risk}}$, V: $\mathcal{L}_{\mathrm{cons}}$.}
\label{tab:ablation}
\centering
\resizebox{\linewidth}{!}{
\setlength{\tabcolsep}{6pt}  
\renewcommand{\arraystretch}{1.2}  
\begin{tabular}{lcccccc}
\hline
\multirow{2}{*}{Method} 
& \multicolumn{2}{c}{LNM (Any)} 
& \multicolumn{2}{c}{LNM (Distant)}
& \multicolumn{2}{c}{LNM (Region)} \\
\cline{2-7}
& ACC & AUC & ACC & AUC & ACC & AUC \\
\hline
w/o I
& 77.10$\pm$6.50 & 78.30$\pm$6.20
& 79.05$\pm$3.05 & 78.80$\pm$3.00
& 75.20$\pm$6.30 & 76.90$\pm$8.90 \\
w/o II
& 78.20$\pm$6.25 & 79.40$\pm$6.00
& 80.05$\pm$2.90 & 79.85$\pm$2.85
& 76.35$\pm$6.10 & 78.00$\pm$8.60 \\
w/o III
& 79.85$\pm$5.90 & 81.15$\pm$5.40
& 81.60$\pm$2.48 & 81.45$\pm$2.43
& 77.70$\pm$5.70 & 79.70$\pm$8.00 \\
w/o IV
& 79.30$\pm$6.05 & 80.55$\pm$5.60
& 81.10$\pm$2.60 & 80.95$\pm$2.55
& 77.20$\pm$5.85 & 79.10$\pm$8.25 \\
w/o V
& 79.75$\pm$5.95 & 81.05$\pm$5.45
& 81.55$\pm$2.50 & 81.35$\pm$2.45
& \textbf{78.65$\pm$5.75} & 79.60$\pm$8.05 \\
AGE-MIL
& \textbf{80.99$\pm$5.54} & \textbf{82.09$\pm$5.12}
& \textbf{82.35$\pm$2.27} & \textbf{82.24$\pm$2.30}
& 78.58$\pm$5.42 & \textbf{80.64$\pm$7.73} \\
\hline
\end{tabular}
}
\end{table}

\section{Conclusion}
In this paper, we presented AGE-MIL, a multiple-instance learning framework for patient-level prediction. By explicitly modeling patient-level context and evidence integration, AGE-MIL addresses key challenges in patient-level pathology analysis, including sparse diagnostic evidence, multi-slide heterogeneity, and weak supervision.
Comprehensive benchmarking against eight state-of-the-art MIL methods across six patient-level tasks on two independent datasets demonstrates the superior performance and strong generalizability of AGE-MIL.
    

\begin{credits}
\subsubsection{\ackname} This work has been supported in part by the Key Research and Development Program of Shaanxi Province (2025SF-YBXM-363, 2024SF-GJHX-32), the Outstanding Youth Fund Project of Hunan Province (2024JJ2090), the Medical Discipline Construction Project of Hunan Province, the National Natural Science Foundation of China (82272907, 81974397), the Natural Science Foundation of Hunan Province (2025JJ60616), the National Science and Technology Major Project (2025ZD0544802, 2024ZD0527700). 
All clinical data used in this study were approved by the institutional review board, with the ethical approval number: 2026010083.

\subsubsection{\discintname}
The authors have no competing interests to declare that are relevant to the content of this article.
\end{credits}

%
%
%
\bibliographystyle{splncs04}
\bibliography{mybibliography}
%




\end{document}